\pdfoutput=1

\documentclass[letterpaper, 10 pt, conference]{ieeeconf}  % Comment this line out if you need a4paper

\IEEEoverridecommandlockouts                              % This command is only needed if 
\usepackage{booktabs}

\usepackage{cite}
\usepackage{amsmath,amssymb,amsfonts}
\usepackage{algorithmic}
\usepackage{graphicx}
\usepackage{textcomp}
\usepackage{xcolor}
\usepackage{subfig}
\usepackage[font=small]{caption}
\usepackage{textcomp}

\usepackage{fancyhdr}
\usepackage[ruled,vlined]{algorithm2e}

\usepackage{amsmath}
\usepackage{fancyhdr,graphicx,amsmath,amssymb}
\usepackage[ruled,vlined]{algorithm2e}
\usepackage{bm}
\usepackage{xcolor}

\usepackage{multirow} % Required for multirows

\DeclareMathOperator*{\argmin}{arg\,min}
\let\labelindent\relax

\include{pythonlisting}

\def\BibTeX{{\rm B\kern-.05em{\sc i\kern-.025em b}\kern-.08em
    T\kern-.1667em\lower.7ex\hbox{E}\kern-.125emX}}

\usepackage{bm} % bold math
\usepackage{color} % change text color        

\usepackage{tikz}
\usetikzlibrary{decorations.pathmorphing} % for snake lines
\usetikzlibrary{matrix} % for block alignment
\usetikzlibrary{arrows} % for arrow heads
\usetikzlibrary{calc} % for manimulation of coordinates
\usepackage{tikzscale}

\tikzstyle{block} = [draw,rectangle,thick,minimum height=2em,minimum width=2em]
\tikzstyle{sum} = [draw,circle,inner sep=0mm,minimum size=2mm]
\tikzstyle{connector} = [->,thick]
\tikzstyle{line} = [thick]
\tikzstyle{branch} = [circle,inner sep=0pt,minimum size=1mm,fill=black,draw=black]
\tikzstyle{guide} = []
\tikzstyle{snakeline} = [connector, decorate, decoration={pre length=0.2cm,
                         post length=0.2cm, snake, amplitude=.4mm,
                         segment length=2mm},thick, magenta, ->]

\usepackage{etoolbox}
\makeatletter
\patchcmd{\@makecaption}
  {\scshape}
  {}
  {}
  {}
\makeatletter
\patchcmd{\@makecaption}
  {\\}
  {.\ }
  {}
  {}
\makeatother
\def\tablename{Table}
\def\tablename{Table}
\usepackage[compact]{titlesec}
\titlespacing{\section}{2pt}{*+1}{*+1}
\titlespacing{\subsection}{2pt}{*+1}{*+1}
\titlespacing{\subsubsection}{2pt}{*+1}{*+1}

\usepackage{url}
\usepackage{hyperref}
\hypersetup{
    colorlinks=true,
    linkcolor=blue,
    filecolor=magenta,      
    }

\usepackage{enumitem}
\setlist{nolistsep,leftmargin=*}
\setlist{nolistsep}
\begin{document}

\title{

% Reactive Whole-Body Obstacle Avoidance for Collision-Free 
% Human-Robot Interaction with Topological Manifold Learning
% % \title{Active Collision Avoidance for Human-Robot Interaction with Topological Manifold Learning
% %\title{Developmentally Constructing Robotic Policy against Dynamic Obstacles and Hazards with Bayesian Reinforcement Learning\\
% %\thanks{Identify applicable funding agency here. If none, delete .}
% 
% SECHROA: Safe and Efficient Collaboration between Humans and Robots in Unstructured Environments with Obstacle Avoidance.

SERA: Safe and Efficient Reactive Obstacle Avoidance for \\ Collaborative Robotic Planning in Unstructured Environments

}

\author{
Apan Dastider 
and
Mingjie Lin
}

%  \author{Albert Author$^{1}$ and Bernard D. Researcher$^{2}$% <-this % stops a space
%  \thanks{*This work was not supported by any organization}% <-this % stops a space
%  \thanks{$^{1}$Albert Author is with Faculty of Electrical Engineering, Mathematics and Computer Science,
%          University of Twente, 7500 AE Enschede, The Netherlands
%          {\tt\small albert.author@papercept.net}}%
%  \thanks{$^{2}$Bernard D. Researcheris with the Department of Electrical Engineering, Wright State University,
%          Dayton, OH 45435, USA
%          {\tt\small b.d.researcher@ieee.org}}%
%  }
%  

% \and

\maketitle

\begin{abstract}

Safe and efficient collaboration among multiple robots in unstructured environments is increasingly critical in the era of Industry 4.0. However, achieving robust and autonomous collaboration among humans and other robots requires modern robotic systems to have effective proximity perception and reactive obstacle avoidance. 
% Unfortunately, most robotic systems operate in dynamically varying environments cluttered with unanticipated obstacles and hazards, making safe collaboration a challenge.
In this paper, we propose a novel methodology for reactive whole-body obstacle avoidance that ensures conflict-free robot-robot interactions even in dynamic environment. Unlike existing approaches based on Jacobian-type, sampling based or geometric techniques, our methodology leverages the latest deep learning advances and topological manifold learning, enabling it to be readily generalized to other problem settings with high computing efficiency and fast graph traversal techniques. Our approach allows a robotic arm to proactively avoid obstacles of arbitrary 3D shapes without direct contact, a significant improvement over traditional industrial cobot settings.

To validate our approach, we implement it on a robotic platform consisting of dual 6-DoF robotic arms with optimized proximity sensor placement, capable of working collaboratively with varying levels of interference. Specifically, one arm performs reactive whole-body obstacle avoidance while achieving its pre-determined objective, while the other arm emulates the presence of a human collaborator with independent and potentially adversarial movements. Our methodology provides a robust and effective solution for safe human-robot collaboration in non-stationary environments.

\end{abstract}


\section{Introduction}
\label{sec:intro}

% homogeneous  policy, compatible state space and action space

\begin{figure}[htbp]
	\centering
	\includegraphics[width=\linewidth]{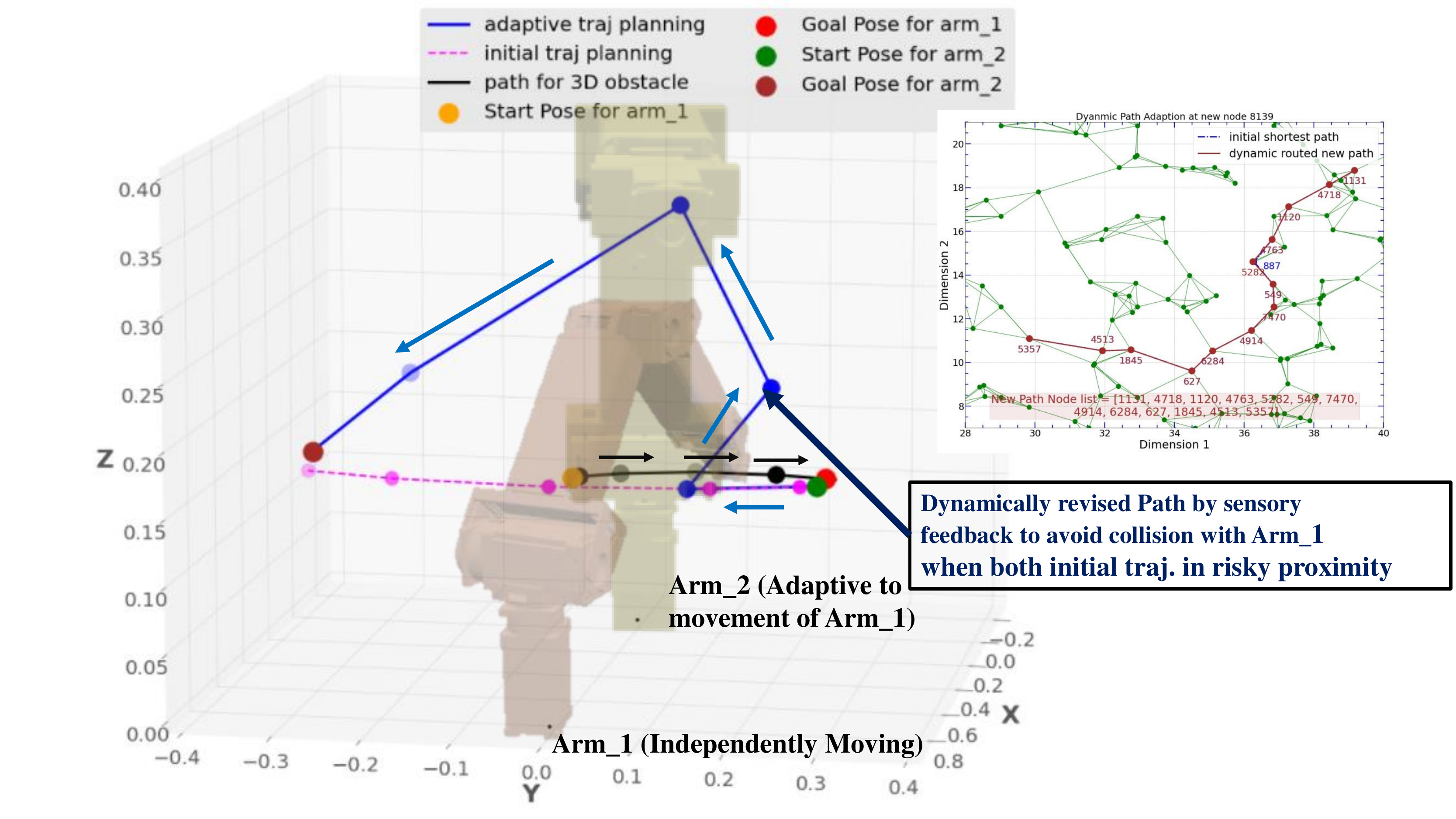}
	\caption{The reactive path planning from sensory feedbacks to avoid probable collisions with arm\_1.}%
	\label{fig:intro}
\end{figure}

%
Effective cooperation between humans and machines or between machines themselves is a defining feature of modern working environments~\cite{Goodrich2007,Schmitt2019a}. However, ensuring conflict-free collaboration and achieving collective goals requires robotic agents to meet two key criteria. Firstly, the agents must be able to adapt to a dynamic working environment, characterized by constantly changing obstacles and system constraints~\cite{Schmitt2019}. Secondly, to ensure timely responses, both human and machine agents must be able to compute an efficient control policy that facilitates real-time learning performance. Addressing these challenges requires an adaptive motion planning algorithm, which is the focus of this paper. Specifically, we present a novel approach to Reactive Obstacle Avoidance named as Safe and Efficient Reactive Obstacle Avoidance (SERA) that enables a robotic agent (\textit{arm\_2}) to expertly avoid a moving 3D obstacle (\textit{arm\_1}), whose form may change over time, while also achieving its own objectives (as illustrated in Figure~\ref{fig:intro}).

In traditional robotic control methods, obstacle avoidance and safe navigation approaches mostly rely on either dynamical system-based approaches\cite{khansari2012dynamical} or sampling based tree search methods\cite{rrt_mani1}. Besides, recent astronomical advancements in deep learning have enabled researchers to deploy various deep reinforcement learning methods to capture intrinsic system dynamics and learn safe obstacle avoidance for complex robotic manipulators\cite{apan}. Beside traditional sampling-based motion planning algorithms such as RRT, RRT*; advance variants like Dynamic RRT*\cite{dynamicrrt} have showed promising performance for handling non-stationary constraints existing in environments and improved RRT have been employed for obstacle avoidance tasks in industrial robotic scenarios\cite{rrt_mani2}. Unfortunately, real-world robotic settings often consist of complex system dynamics and high-dimensional state-space which turns the conventional planning algorithms computationally inefficient for doing real-time decision making in dynamic environments.

To address the challenges of high-dimensional robotic state-space and enable efficient motion planning and skill learning, much attention has been devoted to learn latent space representation and exploit the latent manifold space. Recent works in this area include a system developed by Mohammadi et al. \cite{MohammadiHANR21} that modifies geodesic paths in a Riemannian manifold to achieve obstacle avoidance for a redundant robotic manipulator, and the Learned Latent Space RRT (L2RRT) introduced by Ichter et al. \cite{ichter2019robot}, which performs sampling-based path planning on a latent space representation of the state-space. Additionally, Motion Planning Networks (MpNet) developed by Qureshi et al. \cite{mpnet} employs adaptive replanning methods for handling non-stationary obstacles in the latent space.
While L2RRT and MpNet have shown promising performance in learning various robotic skills in low-dimensional manifold space, our proposed method, SERA, is a more scalable and simpler approach for real-time 3D obstacle avoidance. Instead of expensive sampling-based motion planning, SERA leverages a pre-computed 2D graph-based motion planning method, which is highly efficient and inexpensive. Our method utilizes a densely connected 2D network of manifold representation, which enables efficient traversal for complex robotic motion planning.

% To alleviate this issue, recently much concentrations have been given for learning latent space representation of high-dimensional robotic state-space and exploiting the latent manifold space for motion planning and learning robotic skills. Recently, \cite{MohammadiHANR21} developed an obstacle avoidance learning system for a redundant robotic manipulator through modifying geodesic paths in Reimannian Manifold. Besides, \cite{ichter2019robot} introduced Learned Latent Space RRT (L2RRT) for performing sampling based path planning on a latent space representation of high dimensional robotic state-space. \cite{mpnet} developed Motion Planning Networks which performs sampling based motion planning in latent space with adaptive replanning method for handling non-stationary obstacles. While L2RRT and MpNet have showcased promising performance by learning various robotic skills in low-dimensional manifold space, our method SERA is more scalable and simpler for real-time 3D moving obstacle avoidance due to its pre-computed 2D graph based motion planning instead of expensive sampling-based motion planning. Our method leverages a densely connected 2D network of manifold representation which promised efficient and inexpensive traversal for complex robotic motion planning.  

% \begin{figure}[htbp]
% 	\centering
% 	%\includegraphics[width=\linewidth]{./plots/dual-arm.pdf}
% 	\includegraphics[width=\linewidth]{./figures/sim3D.pdf}
% 	\caption{The experimental setup for our DRL study.}%
% 	\label{fig:intro}
% \end{figure}

Our methodology relies on two fundamental technologies: 1) topological manifold learning and 2) deep neural networks (DNNs) for dimension reduction. In practice, our methodology consists of three key steps.
Firstly, we solve an obstacle avoidance problem by assuming a fixed obstacle location and arbitrary objective. This initial solution serves as the foundation for our overall strategy.
Secondly, we devise an approach to optimally place vicinity sensors on the robotic surface, maximizing detection coverage while minimizing the total number of sensors required.
Finally, we introduce a novel scheme to dynamically determine the most critical collision point when a robotic arm interacts with a 3D obstacle that not only moves, but may also change its form. This dynamic approach enables the robotic agent to effectively respond to changing environmental conditions in real-time, while continuing to satisfy its objectives.

{\bf Contributions}.
We propose a novel approach that integrates topological manifold learning with deep autoencoding to efficiently generate adaptable motion policies for dynamically constrained environments with unforeseen obstacles. Our paper offers the following contributions:

\begin{enumerate}

\item 
Our algorithm leverages manifold learning to embed complex system dynamics in a low-dimensional space, streamlining robotic motion planning. This approach stands out from most obstacle avoidance algorithms, which typically require bespoke solutions for each specific problem setting.

\item 
Our proposed algorithm can adapt to dynamic variations in objectives and working environments in the presence of obstacles or hazards. These obstacles may change their shapes or move arbitrarily, offering significant flexibility. A densely connected network spanning over manifold representation allows for real-time path planning and adaptive navigation to achieve collective objectives. 

\item 
Unlike previous studies, we do not assume the availability of a global camera to detect and localize obstacles or the pose of the robotic manipulator. Instead, we use multiple vicinity sensors strategically placed on the external surface of the robotic arm to localize all relevant objects. 

\end{enumerate}

Taken together, our approach offers an innovative solution to generating adaptable motion policies that can accommodate dynamically changing environments and unforeseen obstacles.

% We propose integrating {\em topological manifold learning} with {\em deep autoencoding}
% synergistically in order to efficiently generate adaptable motion policy
% for dynamically constrained environment with unforeseen obstacles.
% Our paper claims the following contributions:
% 
% \begin{enumerate}
% 
% \item 
% % Deviating from all existing Bayesian RL studies, 
% Our developed algorithm 
% exploits manifold learning 
% to embed complex system dynamics with low-dimensional space in order to 
% drastically simplify robotic motion planning, 
% whereas most existing whole-body obstacle avoidance algorithms need to fully consider 
% specific problems setting individually.
% 
% \item Our proposed  algorithm can generalize to tackle dynamic variations in
% objectives and working environment in the presence of obstacles or hazards.
% In fact, these obstacles can not only dynamically change their shapes but also take arbitrary 
% moving trajectories, therefore 
% providing significant flexibility.
% 
% \item Unlike most prior work, our study doesn't assume a global camera in order
%     to detect and localize all obstacles and the pose of robotic manipulator.
%         Instead, we localize all relevant objects using only multiple vicinity
%         sensors strategically placed on the external surface of our robotic
%         arm.
% 
% 
% \end{enumerate}

\section{Problem Formulation}

% To formulate our work formally, 
% we conceive two robotic agents $arm\_1$ and $arm\_2$ working collaboratively as in Fig.~\ref{fig:hardware}, 
% both equipped with 
% independent trajectory planning respectively.  
% It should be noted that trajectory plannings only need to be functionally compatible, 
% % as defined in Section~\ref{sec:method}, 
% but not necessarily identical.
% Furthermore, we assume the robot arm $arm\_1$ plays a key role for achieving their collective objective, 
% therefore given a prioritized treatment over the agent $arm\_2$.
% As such, $arm\_1$ will act according to its own planned motion  while being oblivious of $arm\_2$.
% However, $arm\_2$, being a subordinate, will pay full attention to $arm\_1$ through receiving sensory feedback.
% Specifically, during their collaborative interaction, $arm\_1$, not intentionally, 
% will dynamically generate
% motion obstacles against $arm\_2$, whereas $arm\_2$ will constantly 
% modify its prior motion planning as depicted in Fig.\ref{fig:intro} in order to flawlessly avoid any hurdles from $arm\_1$. 

In this work, we consider a collaborative setting where two robotic agents, denoted as arm\_1 and arm\_2, are equipped with independent trajectory planning. We assume that the agents share a common objective, with arm\_1 playing a key role in achieving it, and therefore it receives prioritized treatment over arm\_2. Specifically, arm\_1 acts according to its own planned motion while being oblivious to arm\_2, which in turn pays full attention to arm\_1 through receiving sensory feedback. During their collaborative interaction, arm\_1 may dynamically generate motion obstacles against arm\_2, which must be flawlessly avoided by the latter. To formalize this setting, we assume that the trajectory plannings of the two agents need to be functionally compatible but not necessarily identical, and we aim to develop methods for achieving robust and efficient coordination between the agents. Fig.~\ref{fig:hardware} provides an illustration of the hardware setup considered in this work.

\section{Proposed Methodology}

\begin{figure}[htbp]                                       
    \centering                                             
    \includegraphics[width=\linewidth]{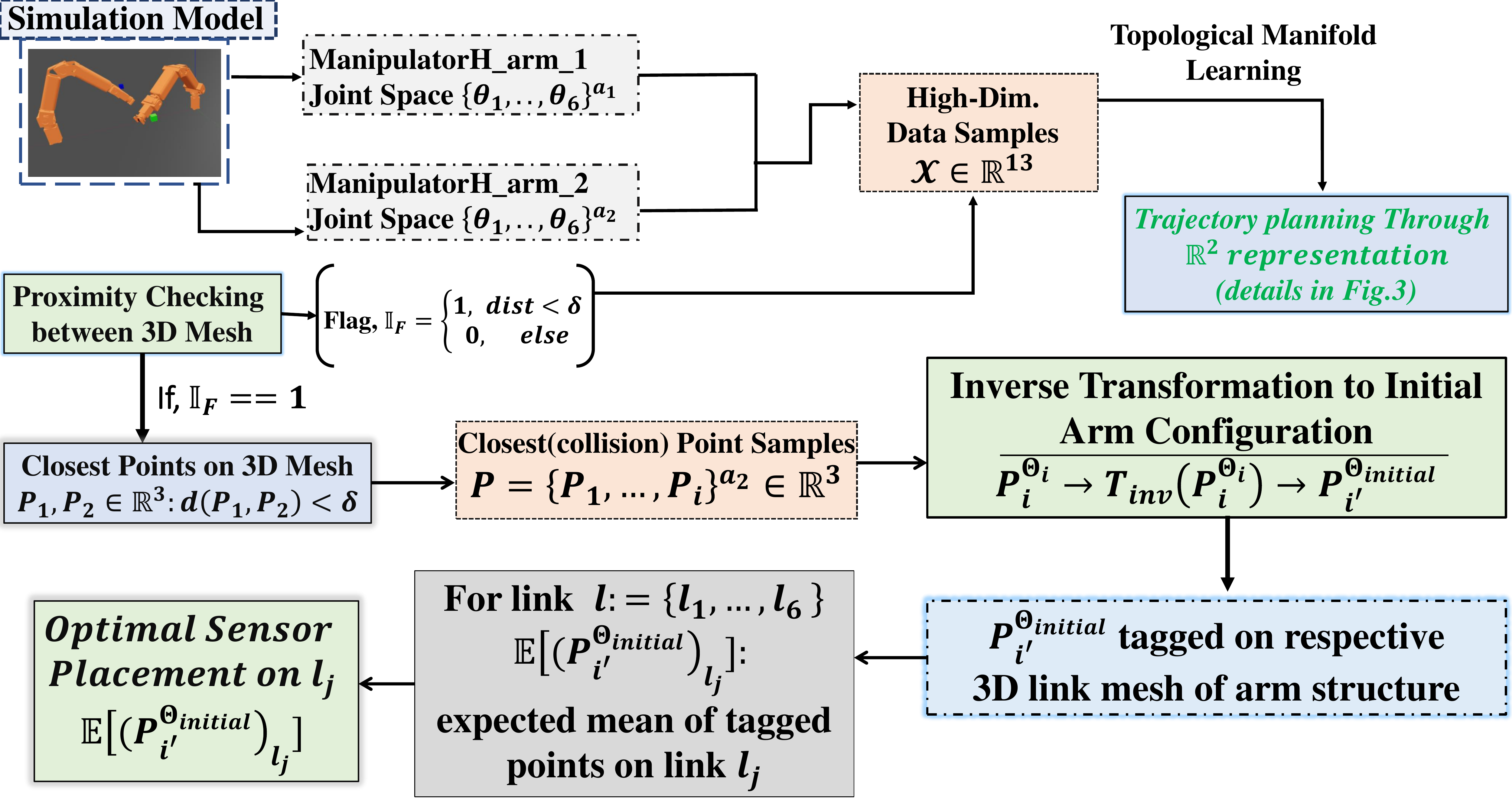}  
    \caption{
        Algorithmic block diagram of Optimal Sensor Placement Method on Different Robotic Links for Efficient Proximity Sensing. 
        }% 
    \label{fig:method}                                   
\end{figure}                                               

% Fig.~\ref{fig:method} depicts the overall structure of our proposed methodology. 
% It consists of two major algorithm modules: 
% %
% 1) optimal proximity sensor placement such that the robotic manipulator can 
% preemptively sense potential collision with moving unknown obstacles.
% 2) manifold-based robotic planning to dynamically avoid moving obstacles, 
% which is accomplished with three sub-modules:  
% a) topological manifold learning, which 
% transforms high-dimensional robotic poses into low-dimensional 2D data points in order 
% to facilitate robotic path planning and obstacle avoidance,  
% b) variational autoencoding/decoding, which bridges between high-dimensional robotic poses 
% and low-dimensional manifold spaces,
% and c)
% graph construction and traversing, 
% which performs robotic control and obstacle avoidance by traversing and rerouting  
% a 2D sparsely-connect graph efficiently.

The proposed methodology is depicted in Fig.~\ref{fig:method}. It consists of two major algorithm modules aimed at ensuring the safety of the robotic manipulator in the presence of moving obstacles.
The first module is the optimal proximity sensor placement algorithm, which is designed to enable the robotic manipulator to preemptively sense potential collisions with unknown obstacles. In this module, we use a commercially available acoustic sensor for ranging based on ultrasound wave propagation. The placement of these sensors is optimized to minimize the number of sensors needed while ensuring adequate coverage.
The second module is the manifold-based robotic planning algorithm, which dynamically avoids moving obstacles. This module is accomplished with three sub-modules. The first sub-module is topological manifold learning, which transforms high-dimensional robotic poses into low-dimensional 2D data points to facilitate robotic path planning and obstacle avoidance. The second sub-module is variational autoencoding/decoding, which bridges the gap between high-dimensional robotic poses and low-dimensional manifold spaces. The third sub-module is graph construction and traversing, which performs robotic control and obstacle avoidance by traversing and rerouting a 2D sparsely-connected graph efficiently. Overall, the proposed methodology ensures that the robotic manipulator operates safely and autonomously in the presence of moving obstacles.

\subsection*{Algorithm Block 1: Optimal Proximity Sensor Placement}
\begin{figure}[htbp]                                       
    \centering                                             
    \includegraphics[width=0.7\linewidth]{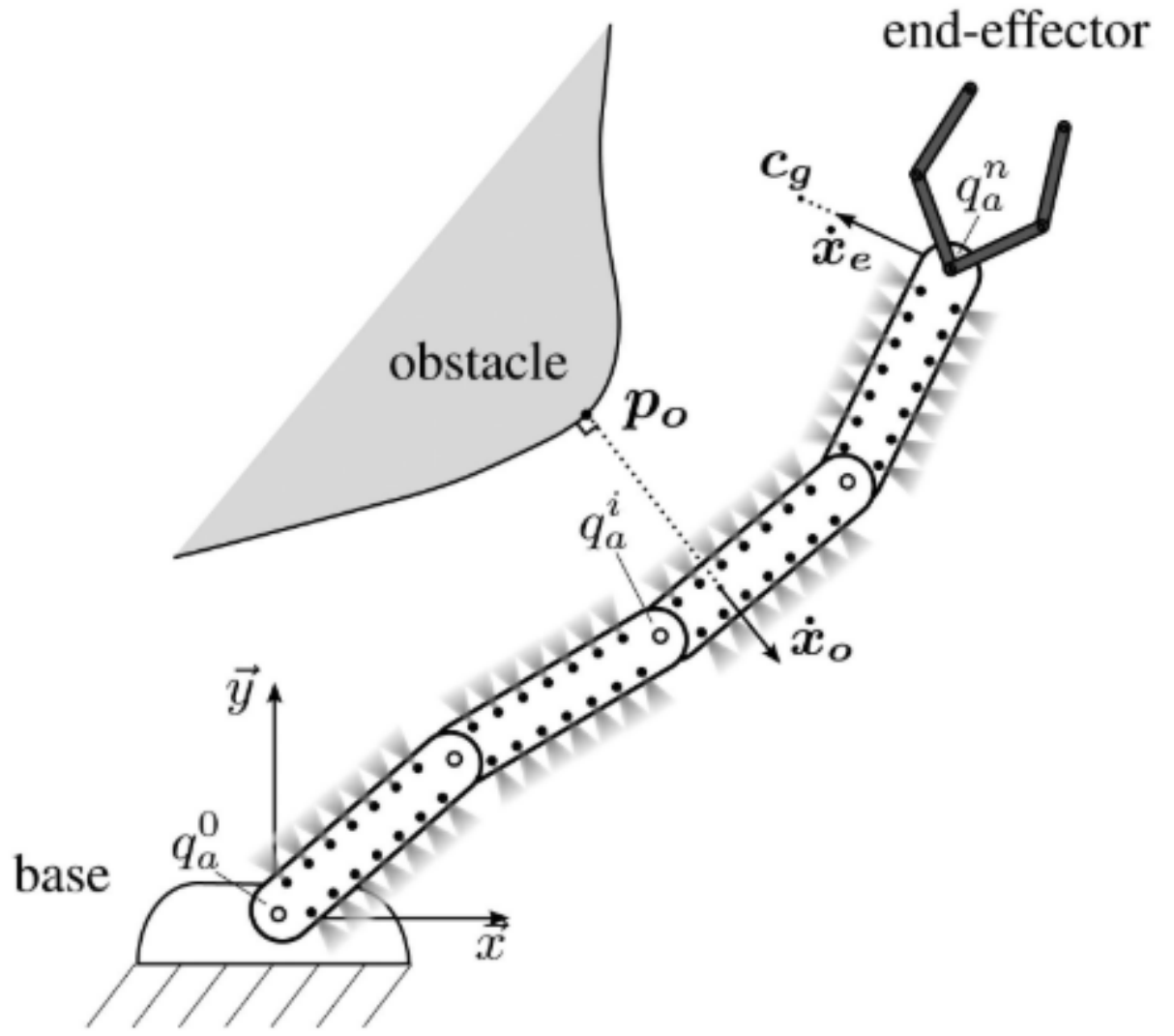}  
    \caption{
        Redundant robot, equipped with proximity sensors on its links, 
        can move toward a target end-effector configuration, 
        while simultaneously moving some of the links of the robot 
        away from an obstacle. (Figure adapted from~\cite{sensor}).
        }% 
    \label{fig:sensor}                                   
\end{figure}                                               

% Proximity perception, 
% central to human-centered robotics,
% can fulfill the promise of safe,
% robust, and autonomous systems in industry and daily life, alongside humans. \cite{sensor1} has provided a detailed review on sensor based control and feedbacks for collaborative robotics.
% %
% In this study, we place all proximity sensors 
% on the exterior of manipulator arms.
% %
% We use commercially available acoustic sensors for ranging based
% ultrasound wave propagation.
% %
% However, unlike in Fig.~\ref{fig:sensor}, we investigate how to optimally place 
% acoustic-based proximity sensors, therefore indirectly minimizing the number of sensors needed. We have accumulated the collision points of both arms from our simulation platform and applied a simple mathematical transformation metric to tag the collision points of robotic arm's 3D collision mesh as shown in Fig.\ref{fig:method}. To find the collision points of 3D mesh on respective link $l_{\{j=2,\cdots,6\}}$, we are required to inversely transform the collision points from world frame to the respective link frame. For that, we have used following inverse relationship between points lying at different coordinate systems,  
% 

Proximity perception is a crucial aspect of human-centered robotics, playing a central role in enabling safe and robust autonomous systems for use in both industry and daily life. A detailed review of sensor-based control and feedback for collaborative robotics can be found in \cite{sensor1}. In this study, we focus on the optimal placement of acoustic-based proximity sensors on the exterior of manipulator arms, with the aim of minimizing the number of sensors required. To achieve this, we first accumulate collision points from both arms in our simulation platform and use a simple mathematical transformation metric to tag the collision points on the robotic arm's 3D collision mesh, as shown in Fig.\ref{fig:method}. In order to find the collision points of the 3D mesh on the respective link $l_{{j=2,\cdots,6}}$, we apply an inverse transformation from the world frame to the respective link frame. This transformation relies on the inverse relationship between points in different coordinate systems, which can be expressed mathematically.

\begin{equation}
    P^{\Theta_j}_{i^{th}\; frame} = [T_0^1...T_{i-1}^i]^{-1}P^{\Theta_j}_{reference\; frame} 
\end{equation}

where, $T_{i-1}^{i} = \left[\begin{array}{c|c c} 
	R_{\theta_{i-1}} & Trans_{i-1}\\ 
	\hline 
	\mathbf{0} & 1 
\end{array}\right]$. 

% Here, $R_{\theta_{i-1}}$ is rotation angle while moving from $(i-1)^{th}$ coordinate frame to frame $i^{th}$ frame and $Trans_{i-1}$ is the respective translation of origin. Since, the simulation tool only gives access to the coordinate values with respect to world frames, we need to transform the collision coordinates to respective link coordinate frames. It assists us to tag the collision points on the correct position of the 3D mesh of accurate link. 
% Moreover, we have also enforced the algorithm to reverse rotations for traversing back to initial pose from any colliding joint configuration after the dataset is fully collected. Thus, we can easily tag the collision points on each respective link for robotic arm.
% Through accumulating a considerable amount of colliding points for each link as well as each plane on the link, we learn a non-uniform distribution over planar co-ordinates for each plane. The expected value $\mathbb{E}[P_x,P_y]$ can be assumed as the optimal proximity sensor location for covering most collision surface.   
% 
To transform the collision coordinates to the respective link coordinate frames, we use the rotation angle $R_{\theta_{i-1}}$ and translation $Trans_{i-1}$ of the $(i-1)^{th}$ coordinate frame to the $i^{th}$ frame. As the simulation tool only provides access to the coordinate values with respect to world frames, this transformation is necessary for accurately tagging collision points on the 3D mesh of each link.
Additionally, we have implemented a reverse rotation algorithm to traverse back to the initial pose from any colliding joint configuration after the dataset is fully collected. This enables us to easily tag the collision points on each respective link for the robotic arm.
By accumulating a significant number of colliding points for each link, as well as for each plane on the link, we are able to learn a non-uniform distribution over planar coordinates for each plane. We can assume the expected value $\mathbb{E}[P_x,P_y]$ as the optimal proximity sensor location for covering most collision surfaces.

\subsection*{Algorithm Block 2: Topological Manifold Learning}

% [fig]
% 
% Manifold learning is a well-known mathematical framework 
% for investigating the geometrical structure of datasets 
% in high-dimensional spaces.
% %
% In this paper, we consider the high-dimensional space 
% defined by $[\{\theta_0, \theta_1, \cdots, \theta_6\}_{arm\_1}; \{\theta_0, \theta_1, \cdots, \theta_6\}_{arm\_2}; \mathbb{I}_F]$,
% where $\{\theta_i\}_{arm\_i}$ determine the exact 
% full-pose of robotic arm $i\in \{1,2\}$ and 
% $\mathbb{I}_F$ is the collision flag.
% %
% Our crucial insight 
% is that the geometrical structure of
% our considered high-dimensional space incoorperates
% the complex system dynamic 
% between  full pose of both robotic arms for arbitrary joint configurations.
% %
% In order to fully extract and leverage the embedding power of our 
% learned manifold,
% we perform a dimension reduction 
% % with the well-known isometric embedding~\cite{Khan2020}
% from high-dimensional robotic space to the low-dimensional manifold 
% space that can be more readily be computed for robotic motion planning.
% %
% It is important to note that all these dimension reductions are done 
% under the condition that the metric tensor is preserved 
% between these two spaces.

Manifold learning is a mathematical framework used to investigate the geometrical structure of datasets in high-dimensional spaces. In this paper, we consider the high-dimensional space defined by the joint configurations of two robotic arms, arm\_1 and arm\_2, denoted by $\{\theta_0, \theta_1, \cdots, \theta_5\}_{arm\_1}$ and $\{\theta_0, \theta_1, \cdots, \theta_5\}_{arm\_2}$ respectively, along with the collision flag $\mathbb{I}_F$. Our crucial insight is that this high-dimensional space incorporates the complex system dynamics between the full poses of both robotic arms for arbitrary joint configurations.

To fully extract and leverage the embedding power of our learned manifold, we perform dimension reduction while preserving the metric tensor between these two spaces. It is important to note that all dimension reductions are performed under this condition to ensure that the geometrical structure of the learned manifold accurately reflects the high-dimensional space of interest.

% the data points are located on a low dimensional manifold embedded in a
% high dimensional Euclidean space. They then try to extract the manifold
% structure and map the data onto the low dimension Euclidean space. These
% methods are sometimes referred to as nonlinear dimension reductions.

\subsection*{Algorithm Block 3: Variational Autoencoding/Decoding}

% [fig]
\begin{figure}[htbp]                                       
    \centering                                             
    \includegraphics[width=\linewidth]{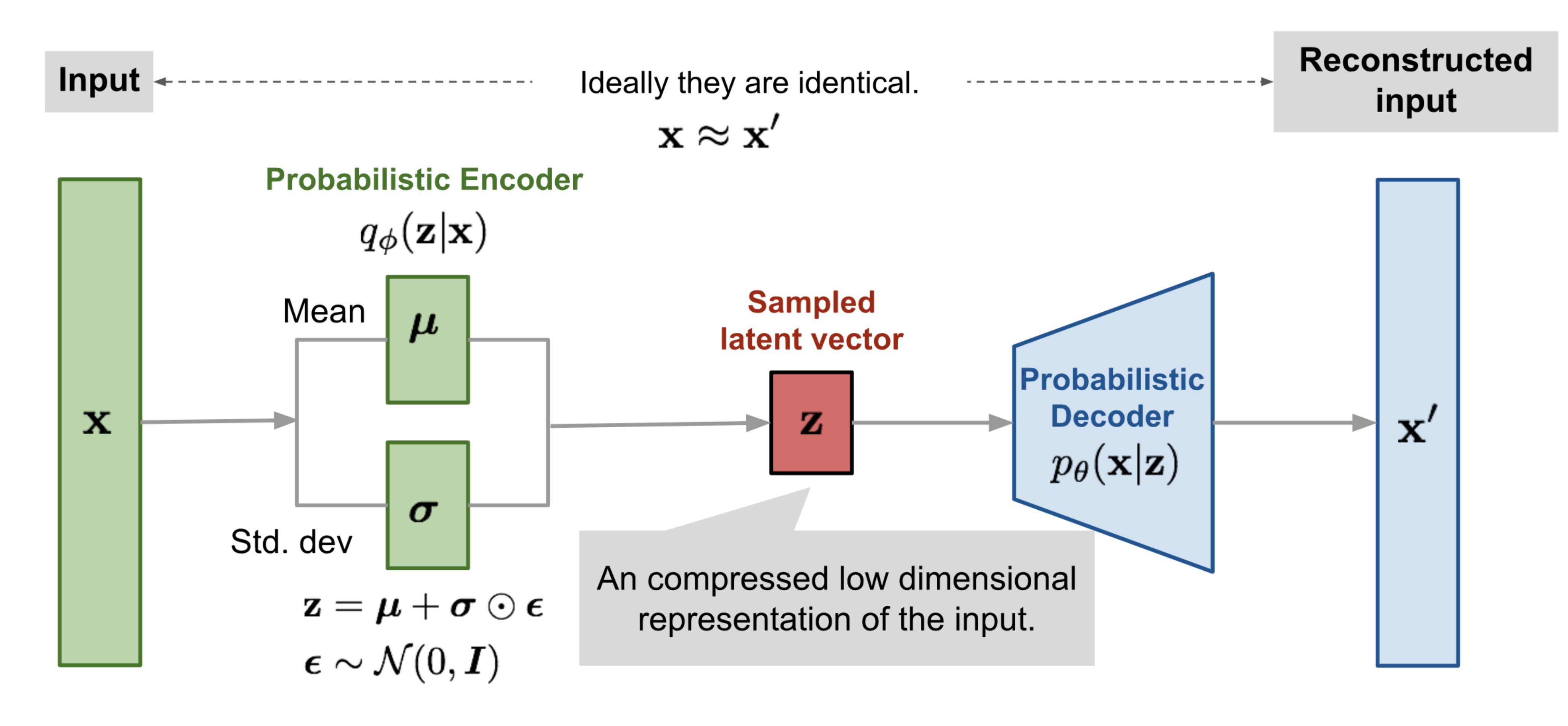}  
    \caption{
        Schematic of a variational autoencoder.
        }% 
    \label{fig:autoencoder}                                   
\end{figure}                                               

To facilitate our manifold-based robotic motion planning, 
we need to seamlessly transform between high-dimensional robotic space $\mathcal{R}$ 
and low-dimensional mathematical space $\mathcal{M}$.
For this, we use a well-established deep learning technique: variational autoencoder (VAE)~\cite{kingma2014autoencoding}.
Conceptually, an autoencoder consists of both an encoder and a decoder.
In fact, 
autoencoders are typically forced to reconstruct the input approximately, 
preserving only the most relevant aspects of the data in the copy.
In our study, we implement autoencoders as deep neural networks
in order to perform dimensionality reduction only.
%
% Schema of a basic Autoencoder
%
Specifically, 
our autoencoder is implemented as  
a feedforward, non-recurrent neural network 
employing an input layer and an output layer connected by one or more hidden layers. The
output layer has the same number of nodes (neurons) as the input layer. 
The purpose of our autoencoder is to reconstruct its inputs (minimizing the difference between the
input and the output) instead of predicting a target value $\mathbf{Y}$
given inputs $\mathbf{X}$.

Our autoencoder consists of two parts, the encoder and the decoder, 
which can be defined as transformations $\phi$  and $\psi$, such that:
$\phi :{\mathcal {R}}\rightarrow {\mathcal {M}}$
and 
$\phi :{\mathcal {M}}\rightarrow {\mathcal {R}}$
and 
$\phi ,\psi =\argmin \|{\mathcal {X}}-(\psi \circ \phi ){\mathcal {X}}\|^{2}$.
Autoencoders are trained to minimise reconstruction errors (such as squared errors), 
often referred to as the ``loss" $\mathcal {L}(\mathbf {x} ,\mathbf {x'} )=\|\mathbf {x} -\mathbf {x'} \|^{2}$,
where $\mathbf {x}$  is usually averaged over the training set.
Our autoencoder is trained through backpropagation of the error.
Conceptually, the feature space of our autoencoder is the low-dimension 
manifold representations produced by manifold learning, 
therefore having lower dimensionality than the input space 
$\mathcal {R}$, which, in our study, is the pose of our dual 6-DoF robotic manipulators.
As such, the feature vector $\phi (x)$ after manifold learning can be
regarded as a compressed representation of the input $x$. 

\begin{figure}[htbp]                                       
    \centering                                             
    \includegraphics[width=\linewidth]{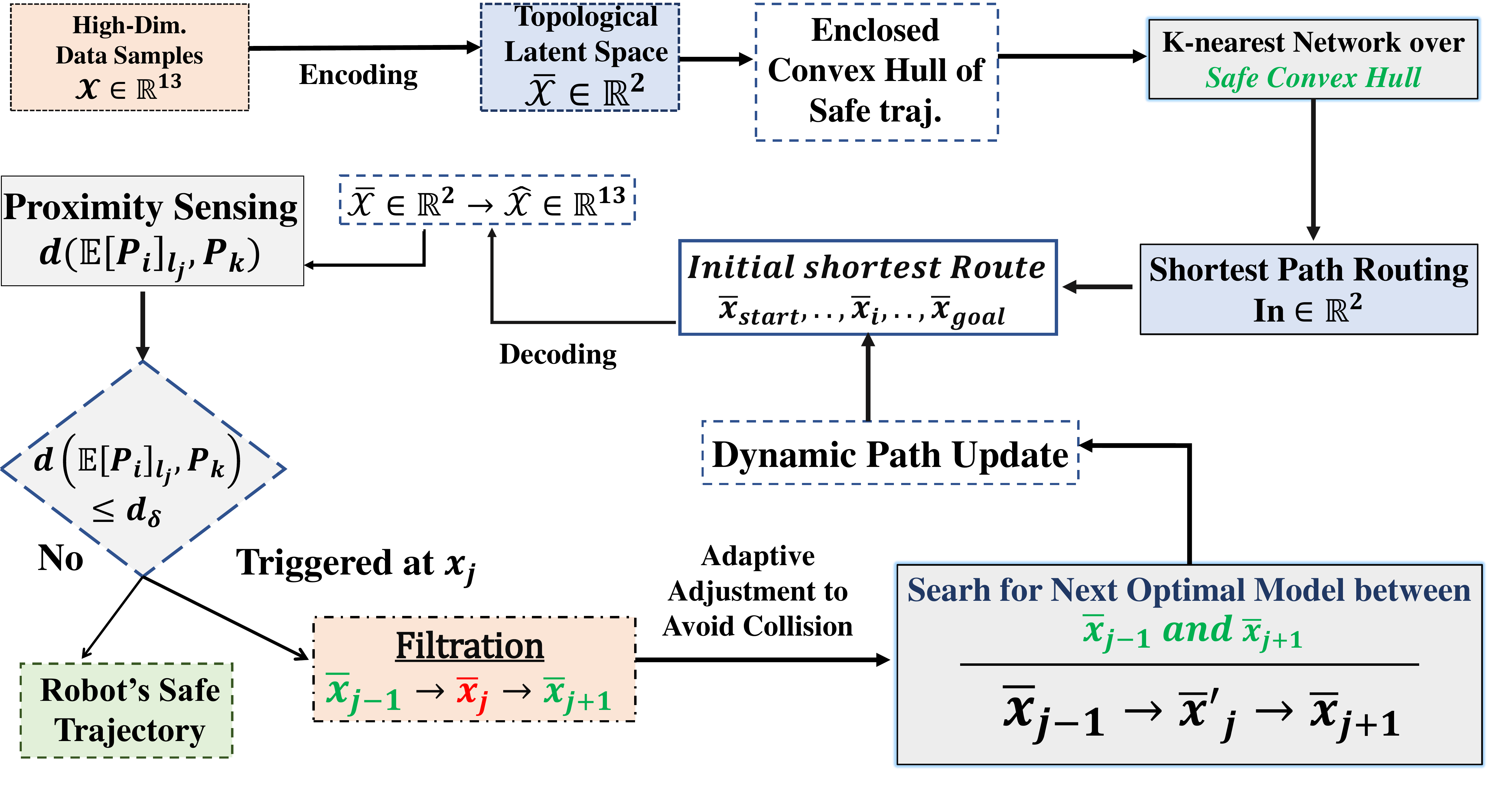}  
    \caption{Block Diagram of Trajectory Planning Through Manifold Space
        }% 
    \label{fig:block2}                                   
\end{figure} 

\subsection*{Algorithm Block 4: Graph Construction and Traversing}

Graph-based motion planning has become an established technique for mobile robots in complex environments \cite{8968151, KupervasserDMRobot}. However, in our study, we extend this approach to robotic manipulators in the context of 3D moving meshes, such as another robotic arm. To capture the system dynamics in this scenario, we construct a topological manifold space graph. To simplify the graph structure and traversal, we choose to use a 2D manifold space. Once the manifold space is created, we label its vertices as either "collision" or "collision-free". During motion planning, we use Dijkstra's algorithm to perform shortest-distance routing, using only the green vertices ("collision-free").
To represent our graph, we implement an adjacency list structure using Python dictionary data structures. Specifically, we use a dictionary of dictionaries, where the outer dictionary is keyed by nodes and the values are dictionaries keyed by neighboring nodes, with associated edge attributes. This "dict-of-dicts" structure allows for fast addition, deletion, and lookup of nodes and neighbors in large graphs. Our software code is based on the open-source NetworkX package \cite{SciPyProceedings_11}, and all of our functions manipulate graph-like objects solely via predefined API methods, rather than acting directly on the data structure. We start with an initial shortest path from a start node to target vertice on the 2D connected network. For each node on the path, we check the proximity between nearest collision meshes of both robotic manipulators. The proximity sensors placed on the optimal positions calculate these distances between closest link meshes. If the proximity distance violates a safety threshold, we easily filter that node from the initial planned path and dynamically update the connecting trajectory with new collision-free node to reach the target location. Thus, our fully connected graph network enables our adaptive robotic arm to avoid collision while achieving preset objective in an efficient way.

\section{System Overview}
\label{sec:system}

\subsection{Experimental Platform and Simulation Setup}

\begin{figure}[htbp]
    \centering
    \includegraphics[width=\linewidth]{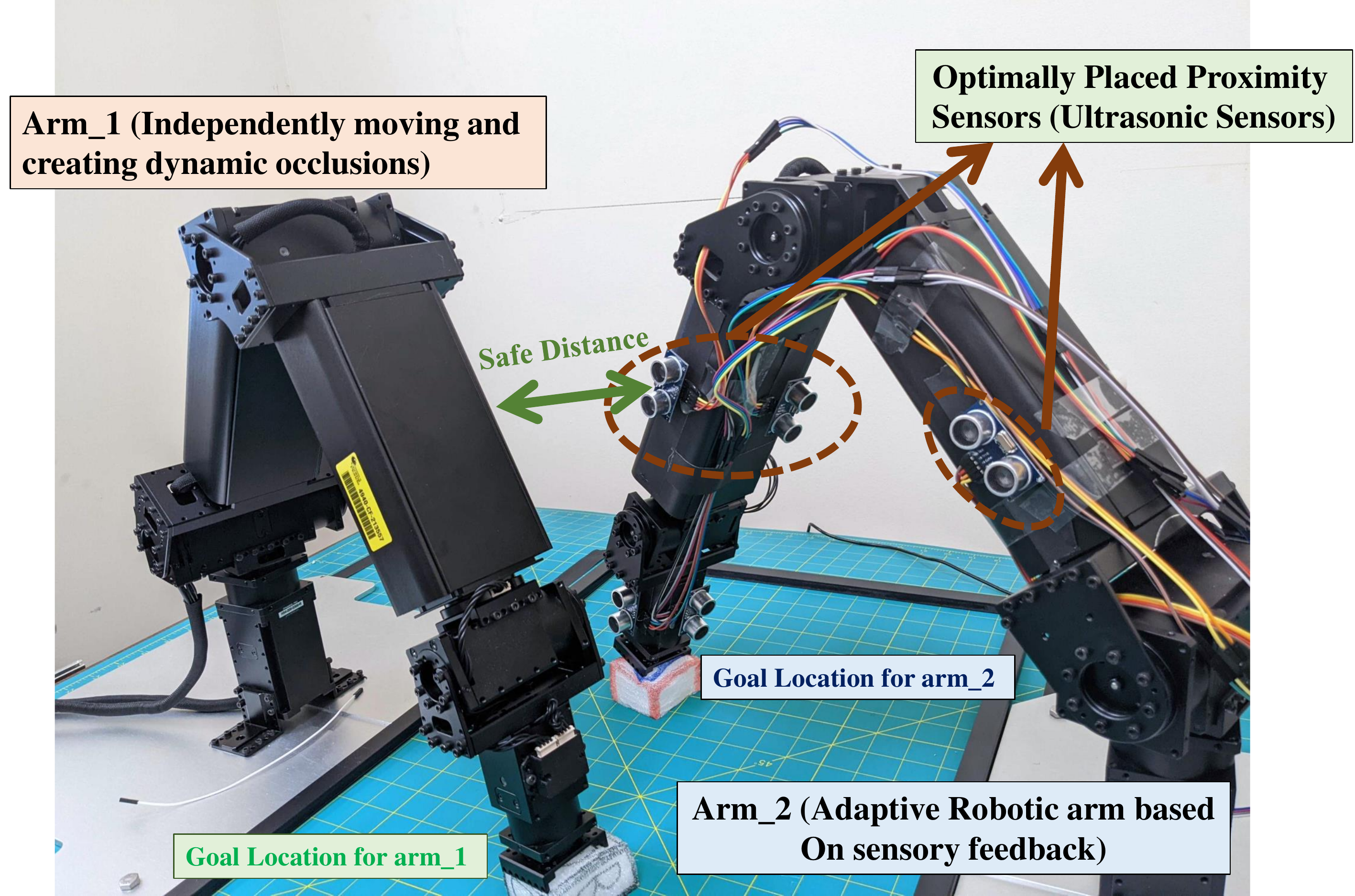}
    \caption{Hardware Platform with Sensor Attachment}%
    \label{fig:hardware}
\end{figure}

Fig.\ref{fig:hardware} shows our experimental setup, which consists of two Robotis{\textregistered} Manipulator-H research-grade 6-DOF robotic manipulators, arm\_1 and arm\_2, mounted on a grid-mapped table-top that serves as a shared and compact workspace for all experiments. We have installed a $400\times600$ resolution touch-sensitive frame to locate target positions for both arms.
arm\_1 is used to create dynamic occlusions while arm\_2 maneuvers. The adaptive planning algorithm runs on a Lambda QUAD GPU workstation with an Intel Core-i9-9820X (10 cores) processor and Ubuntu 18.04. arm\_2 is equipped with proximity sensors optimally located for accurate collision detection and adaptive route traversing. The sensors' feedback is recorded through a Raspberry PI to the controller. Our mechanical platform is integrated with PID values to introduce precise position control \cite{Ota2019} within a safe range.

% As shown in Fig.\ref{fig:hardware},
% we used here two research-grade 6-DOF robotic manipulators of
% Robotis{\textregistered} Manipulator-H,
% $arm\_1$ and $arm\_2$, that are
% mounted on 
% a grid mapped table-top which served as
% a shared and concise workspace  for all experiments. We have installed a $400\times600$ resolution touch-sensitive frame to locate the target positions for both arms. Specifically, $arm\_1$ serves the purpose of creating dynamic occlusions while maneuvering for the adaptive robotic agent $arm\_2$. The adaptive planning algorithm runs on a Lambda QUAD GPU workstation equipped with Intel Core-i9-9820X (10 cores) processor and Ubuntu 18.04. Moreover, the adaptive entity $arm\_2$ is empowered with proximity sensors embodied at optimal location for accurate collision detection followed by adaptive route traversing. The feedback of these sensors are recorded through a Raspberry PI to the controller. Our mechanical platform is integrated with PID values to introduce  precise position control \cite{Ota2019} within safe range. 
% 

To replicate our hardware setup and validate our approach, we created high-fidelity robot models using Robotics Toolbox for Python \cite{rtb}. We used URDF files to easily replicate the models inside the simulated environment and exploited the available functionalities to check collisions among 3D meshes or tag proximal points among 3D meshes in the simulation. To ensure low-latency data processing and parallel execution among simulation environments and real hardware, we encapsulated the framework inside the Robot Operating System (ROS) ecosystem and established an inter-processing communication (IPC) protocol. This allowed us to accurately simulate the behavior of our real hardware setup and gather the necessary data to train our algorithms.
%On the other hand, to replicate the hardware setup in order to accumulate dataset and validate our approach, we have created high-fidelity robot models inside Robotics Toolbox for Python \cite{rtb} which has been very recently forked out of its older version well-known for Matlab. With the open-sourced robot description files--URDF files, we easily replicated the models inside the simulated environment and we exploited the available functionalities for checking collision among 3D meshes or tagging proximal points among 3D meshes in the simulation. For establishing the inter-processing communication (IPC) protocol to ensure low-latency data processing and parallel execution among simulation environments and real hardware, we have encapsulated the framework inside the Robot Operating System (ROS) ecosystem.    

\begin{figure*}[htbp]
    \includegraphics[width=1\textwidth]{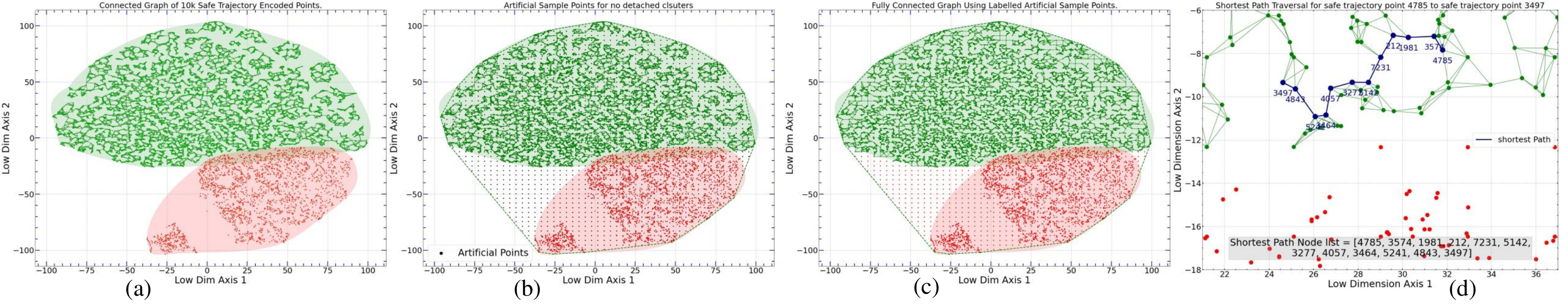}
    \caption{(a) Connected Network over 10k Manifold Points, (b)-(c) Uniformly Sampled Points for a densely Connected Network, (d) Shortest Path Routing by Dijkstra's algorithm }
    \label{fig:shortest}
\end{figure*}

\subsection{Experimental Procedures and Learning Variables}

The text describes key variables in high-dimensional data used to build a latent space manifold representation. The dataset includes two sets of six joint angles, 
% each set containing angles (${\theta_i^a}{i=0,..,5}^{a=1,2}$), 
and a binary collision flag ($\mathbb{I_F}$). Additionally, coordinates of the closest points on 3-D meshes while colliding with links $\{l_i\}_{i=2,..,6}$ of arm\_2 were logged to optimally allocate proximity sensors. The control command $\Theta_i^a$ for each arm is a vector of 6 joint angles. The data accumulation process was stochastic, and the simulation platform was used for dataset collection and initial learning. The VAE architecture used Pytorch and comprised a decoder and encoder network with three layers of neurons ({450, 250, 100}). The encoded latent space representation is in $\mathbb{R}^2$, which facilitates graph embedding for planning optimal paths from node to node. Validation experiments were conducted using a real hardware setup with varying goal locations and random occlusions created by arm\_1.

% We now define key variables in high-dimensional data for building the latent space manifold representation.
% % 
%  The dataset comprises 13 variables-6x2 joint angles $\{\theta_i^a\}_{i=0,..,5}^{a=1,2}$, and binary collision flag,  $\mathbb{C_F}$. Besides, to optimally allocate the proximity sensors on different links $\{l_i\}_{i=2,..,6}$ of $arm\_2$, we also logged the closest point coordinates with link IDs on 3-D meshes while colliding. Since, base link is fixed and out of the confined workspace, it is unnecessary to tag collision points on it. The control command $\Theta_i^a$ for each arm is a vector of 
% 6 joint angles. Since, data accumulation process is very stochastic in nature and both robotic arms can collide with each other for multiple instances, we used the simulation platform for dataset collection and initial learning purposes. When the algorithm and manifold construction converges, the validation experiments are carried over varying goal locations and random occlusions created by $arm\_1$ in real hardware setup for real-time and scalable applications of our algorithms. Our VAE architecture is implemented on Pytorch \cite{Pytorch}. The decoder and encoder network comprises on three layers with \{450, 250. 100\} neurons. The encoded latent space representation lies in $\mathbb{R}^2$ which greatly facilitates creating graph embedding to plan optimal path from node to node. 
% 

\begin{figure}[htbp]
    \includegraphics[width=\linewidth]{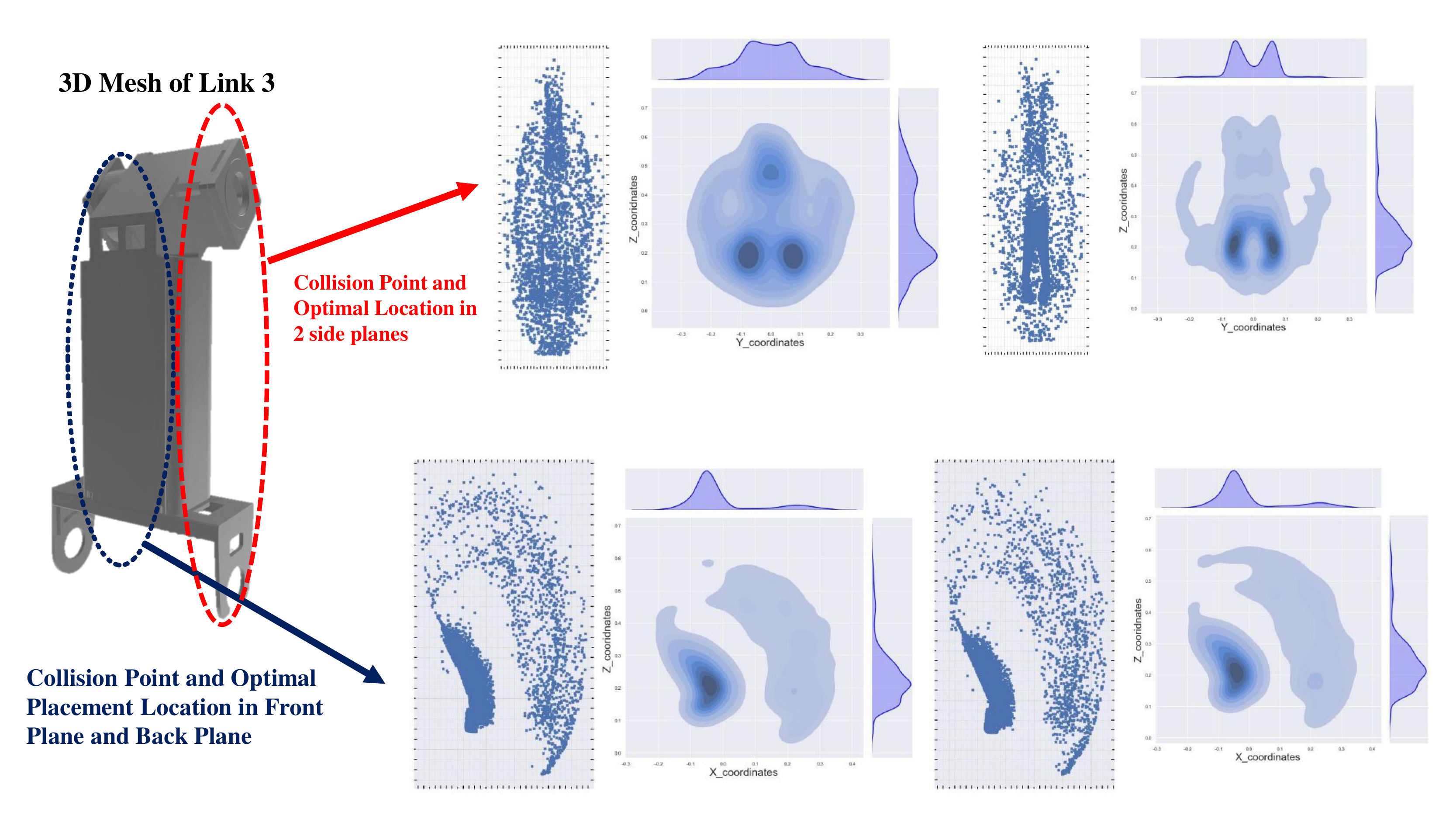}
    \caption{Collision points on link 3 and Optimal sensor placement location calculated through non uniform distributions}
    \label{collision}
\end{figure}

\begin{figure}[htbp]
    \includegraphics[width=\linewidth]{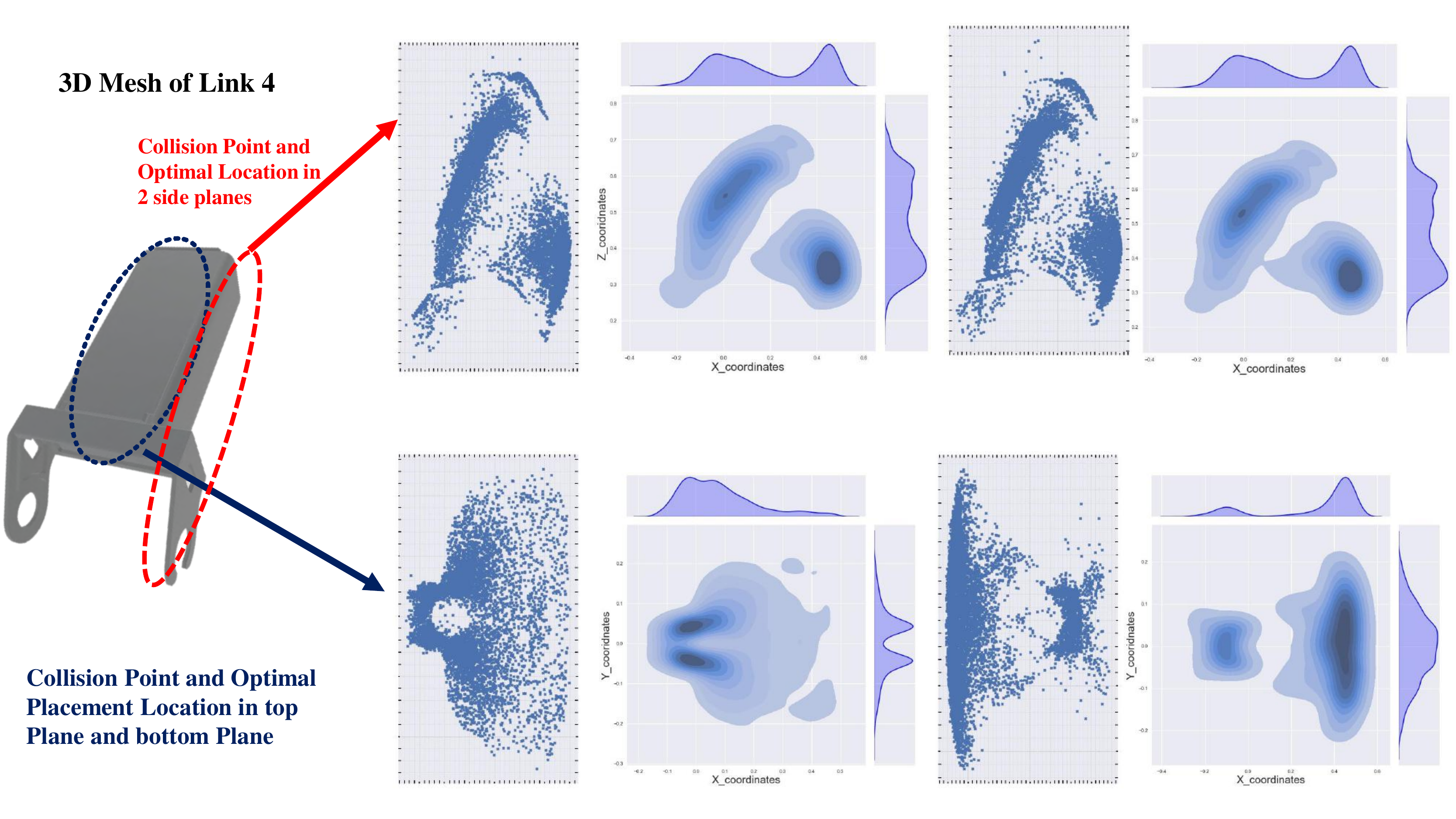}
    \caption{Collision points on link 4 and Optimal sensor placement location calculated through non uniform distributions}
    \label{l4}
\end{figure}

\begin{figure}[htbp]
    \includegraphics[width=\linewidth]{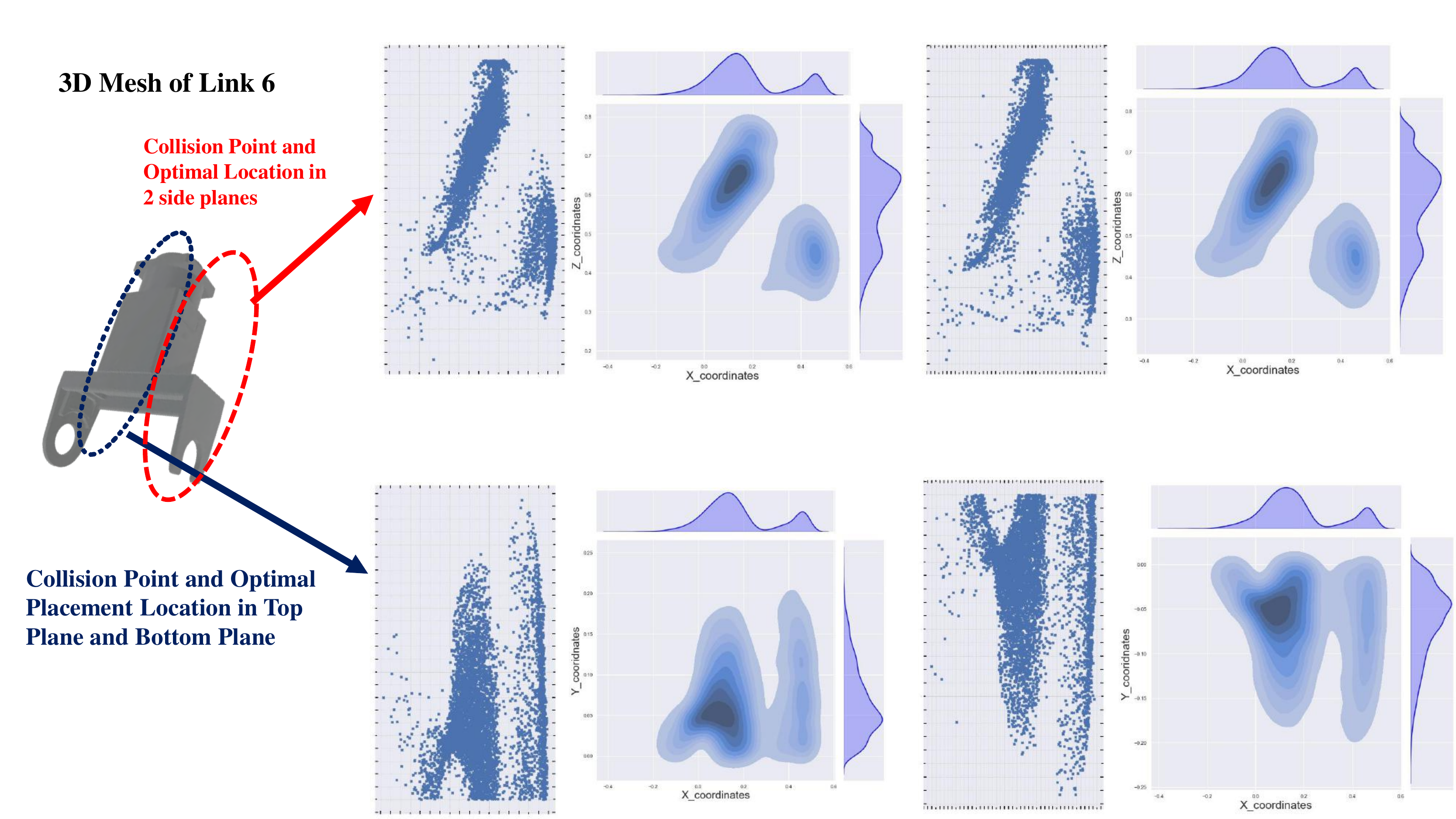}
    \caption{Collision points on link 6 and Optimal sensor placement location calculated through non uniform distributions}
    \label{l6}
\end{figure}

\section{Results Analysis}
Our experiments seek to investigate the following: 

\begin{enumerate}

\item Can we create a resolution to optimal sensor placement problem through tagging the colliding points to respective links and creating the non-uniform distributions over coordinate values for each plane of cuboid links?

\item Can we learn two disparate convex hulls in $\mathbb{R}^2$ space representing the safe and colliding samples for initial feasible trajectory?  

\item Can we concurrently handle the unseen perturbations and adaptively re-plan trajectory over \textit{safe} convex hull to reach the goal location?  

\end{enumerate}

%%%%%%%%%%%%%%%%%%%%%%%%%%%%%%%%%%%%%%%%%%%%%%%
\subsection{Ideal and Efficient Sensor Placement}

% In this work, we attempted to address the issue of positioning the proximity sensors optimally on mechanical structure of the different links. 
To ensure effective collision avoidance for our adaptive robotic arm, proximity sensors need to be placed in locations with a high probability of collision with the robot structure. To address this issue, our dataset includes the closest collision points in 3D space on the robotic arm's collision meshes. By applying the inverse transformation matrix discussed in Section III and the inverse rotation to the initial pose from the robot's $\Theta^j$, we can tag these collision points on the mechanical structure of the robotic arm. Optimal sensor positions for the four different planes of the three longest links of the Manipulator-H robotic arm are shown in Figures \ref{collision}, \ref{l4}, and \ref{l6}. The joint plots in these figures are based on coordinate values from the respective planes of the links when the robot is spawned at its initial locations.
The estimated distributions of independent variables in these plots reveal the high-probability collision locations, and sensors should be placed at these points to provide effective feedback for collision avoidance. For example, in the top plane of Link 6, which is parallel to the $X-Y$ plane, only the $\{P_{x_j},P_{y_j}\}_{l_6}$ collision coordinates are used to show the optimal sensor placement locations. As described in the methodology section, the mean value $\mathbb{E}[P{x_j},P_{y_j}]$ is assumed to be the optimal sensor placement location on the top plane of Link 6 for the most efficient collision avoidance feedback and dynamic re-planning to ensure a safe trajectory.
Overall, Figures \ref{collision}, \ref{l4}, and \ref{l6} show the most effective and systematic sensor placement locations for the significant and longest links of our adaptive manipulator, enabling efficient collision avoidance and safer trajectory planning.

\begin{figure*}[htbp]
    \includegraphics[width=1\textwidth]{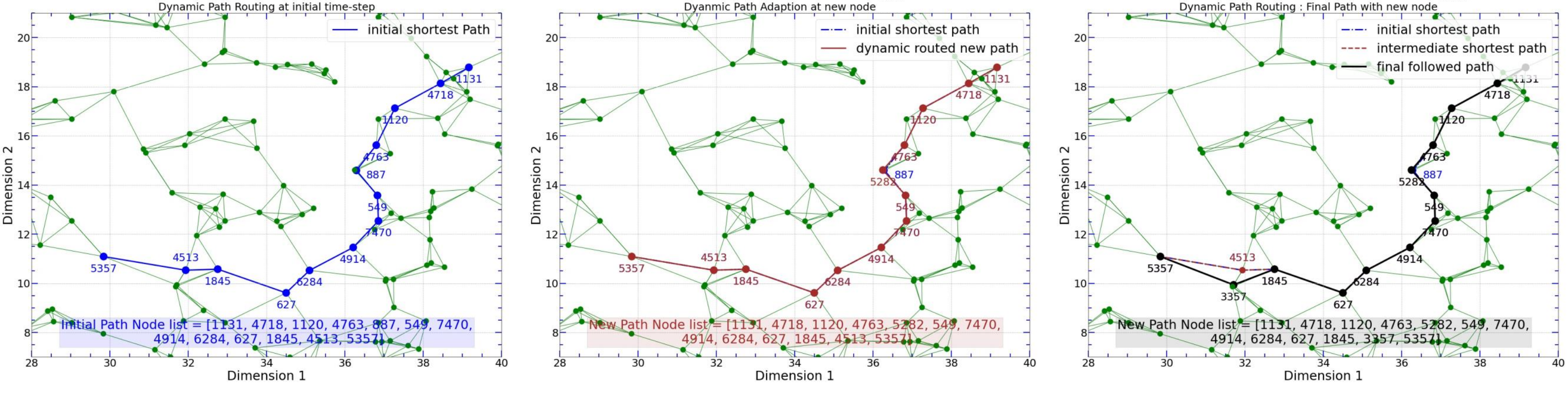}
    \caption{Dynamic Path routing for avoiding collision: (a) Initial Sampled Path (b) Path Change for avoiding first perturbation (c) Comparison of final followed path and initial path }
    \label{fig:adaptive}
\end{figure*}

\begin{figure*}[htbp]
    \includegraphics[width=1\textwidth]{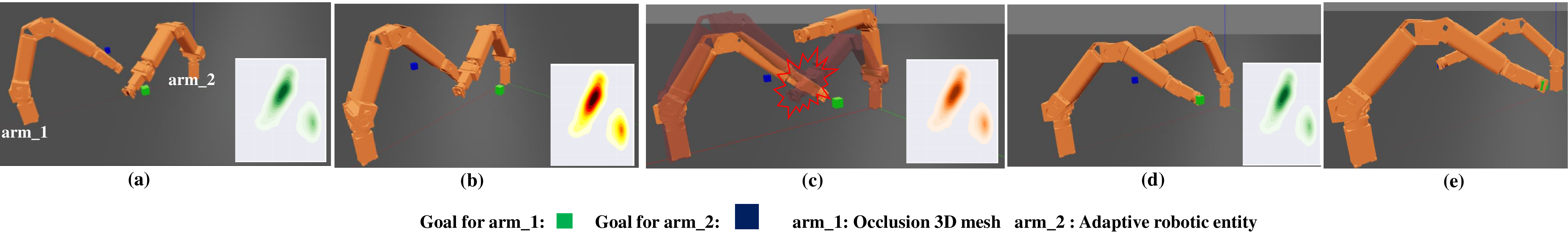}
    \caption{Dynamic Trajectory Planning: (a) Initial Pose (b) Both arms on verge of collision--sensors triggered (c) adaptive $arm\_2$ re-routes through new node and adapts new trajectory to avoid collision--sensors feedback slightly changes (d) both arms following safe trajectory--feedback neutralises}
    \label{fig:sim}
\end{figure*}

\subsection{Latent Space Representation and Routing over Graphs}
\begin{figure}[htbp]
    \centering
    \includegraphics[width=\linewidth]{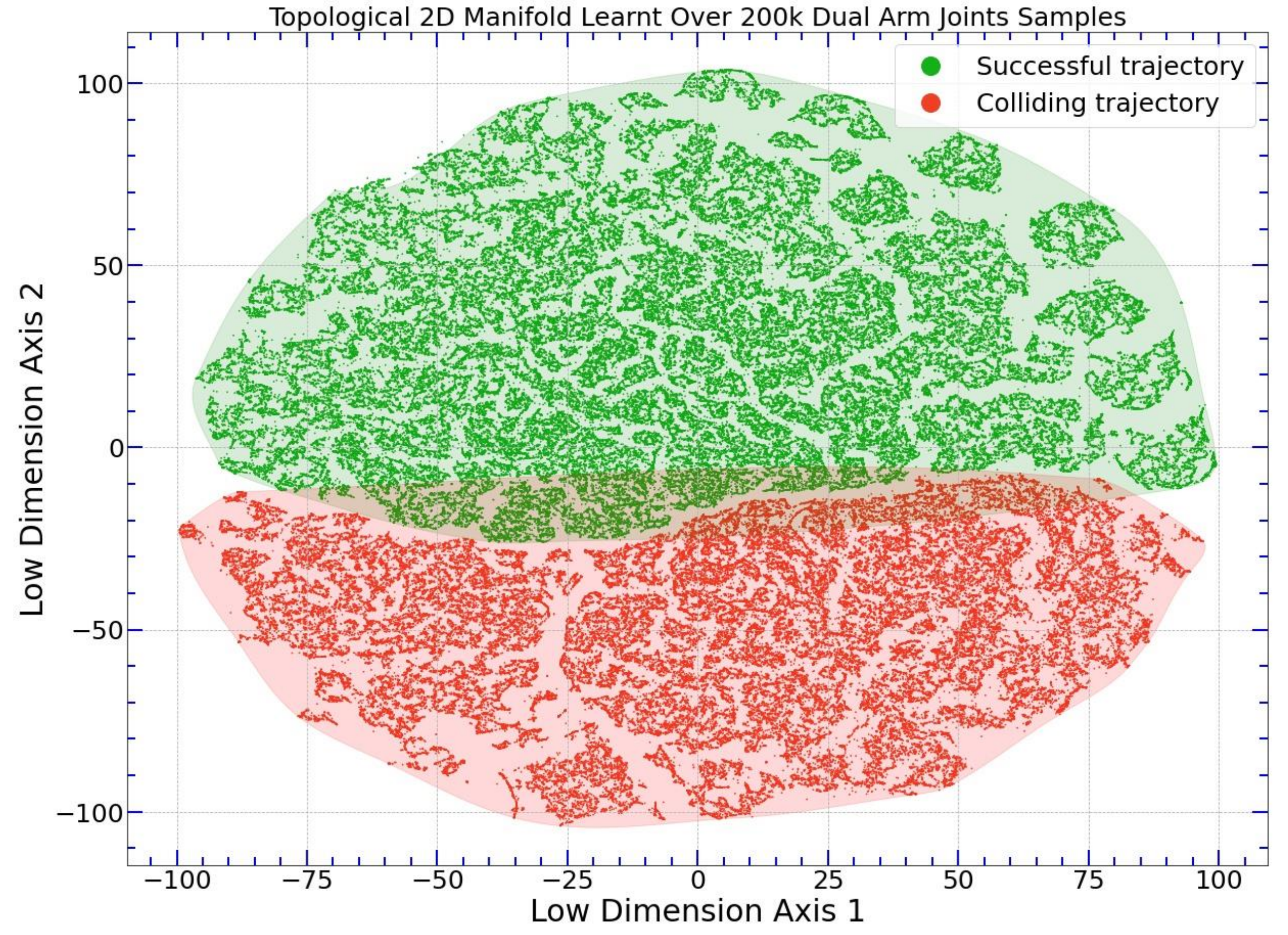}
    \caption{Two separate convex hulls}%
    \label{fig:hulls}
\end{figure}

Each data sample comprises the joint angle values from both arms as well as a collision flag $I_F$ which is either $0$ when collided or $1$ when a safe trajectory is finished. With this high-dimensional representation, our VAE learns a topological manifold representation in $\mathbb{R}^2$. As shown in Fig.\ref{fig:hulls}, the latent space manifold sample points can be smoothly separated between two clusters of safe samples and collision samples. 
The collision indicator flag and inter-relationship between synchronously sampled joint space poses for safe trajectories and colliding trajectories enable the algorithm to effectively learn these disparate manifold representation through variational encoding.      
In the next step of our research, we create a connected graph solely over the safe samples space, which is highlighted in green in Fig. \ref{fig:hulls}. To achieve this, we have employed an unsupervised K-Nearest Neighbor algorithm to connect neighboring vertices in the safe manifold space. The resulting connected graph network allows for the shortest path routing from any vertex as a start pose to another vertex as a goal position for the adaptive robotic agent.
As shown in Fig. \ref{fig:shortest}(a), sub-clusters are automatically generated within the safe manifold while creating the connected graphs. However, this generation of disparate sub-clusters inside safe samples poses a challenge to the notion of complete path routing from any random node to another random node on the connected network. To ensure a dense connection, we artificially sampled points in the same $\mathbb{R}^2$ space and used our trained decoder part of VAE to classify these points as safe or colliding trajectories, as shown in Fig. \ref{fig:shortest}(c). The similarity of labels between Fig. \ref{fig:hulls} and Fig. \ref{fig:shortest}(c) confirms the successful preservation of the global structure of the manifold space through Variational Encoding.
We now have a densely connected network over all green points, which guarantees that our initial routed path can be considered a safer trajectory. To determine the shortest path, we utilized Dijkstra's algorithm, a well-known shortest routing algorithm over connected networks.

\begin{table}[]
\centering
\begin{tabular}{|ccccc|}
\hline
\multicolumn{1}{|c|}{\multirow{3}{*}{Methods}}                                                  & \multicolumn{4}{c|}{Mode of Operation}                                                                                                                                                                                                                                                               \\ \cline{2-5} 
\multicolumn{1}{|c|}{}                                                                          & \multicolumn{2}{c|}{Total Runtime, T (s)}                                                                                                               & \multicolumn{2}{c|}{Success Ratio (\%SR)}                                                                                                  \\ \cline{2-5} 
\multicolumn{1}{|c|}{}                                                                          & \multicolumn{1}{c|}{A}                                                     & \multicolumn{1}{c|}{B}                                                     & \multicolumn{1}{c|}{A}                                                          & B                                                        \\ \hline \hline
\multicolumn{1}{|c|}{SERA}                                                                       & \multicolumn{1}{c|}{5.3}                                                   & \multicolumn{1}{c|}{9.1}                                                   & \multicolumn{1}{c|}{99.6}                                                       & 95.3                                                     \\ \hline
\multicolumn{1}{|c|}{DynamicRRT*\cite{dynamicrrt}}                                                               & \multicolumn{1}{c|}{52.6}                                                  & \multicolumn{1}{c|}{68.1}                                                  & \multicolumn{1}{c|}{79.3}                                                       & 66.7                                                     \\ \hline
\multicolumn{1}{|c|}{MpNet\cite{mpnet}}                                                                     & \multicolumn{1}{c|}{22.6}                                                  & \multicolumn{1}{c|}{25.4}                                                  & \multicolumn{1}{c|}{85.1}                                                       & 82.5                                                     \\ \hline \hline
\multicolumn{5}{|c|}{Mode of Operation Details:}                                                                                                                                                                                                                                                                                                                                                       \\ \hline
\multicolumn{5}{|l|}{\begin{tabular}[c]{@{}l@{}}A : Simple level of complexity : Both arm need to reach own single \\        preset target location\\  B : Hard level of complexity : Multiple target locations are set \\        sequentially to reach by the robotic arms while avoiding \\        collision, required to multi-level path re-planning in \\        intermediate nodes\end{tabular}} \\ \hline
\end{tabular}
\caption{Comparison With two baseline approach}
\label{Table1}
\end{table}

\begin{figure*}[htbp]
    \includegraphics[width=1\textwidth]{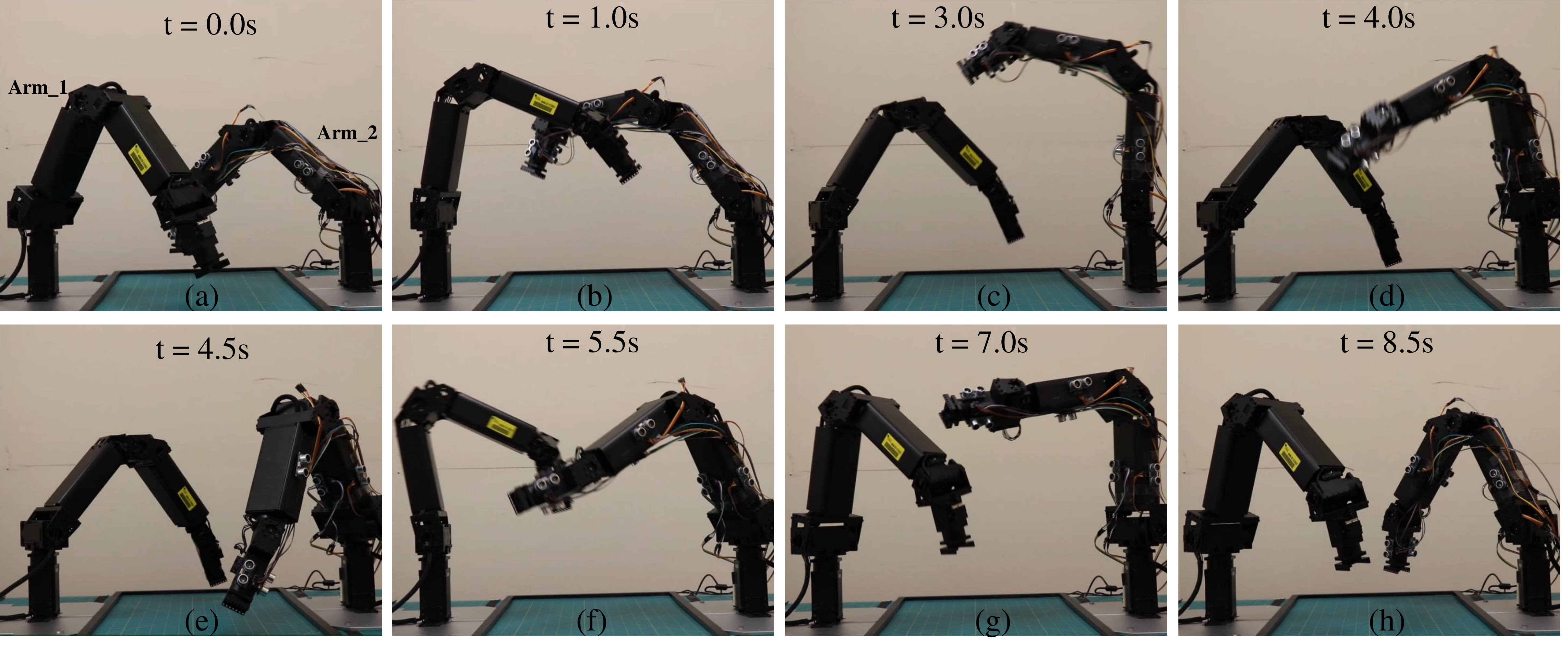}
    \caption{(a) Initial Pose, where arm\_1 creates dynamic obstacles and arm\_2 is the adaptive learner (b) Both arms reach for preset target location, but probable collision can happen when both arms get close to each other.(c)-(d) Adaptive Learning agent plans through a new node on manifold network to avoid probable collision with arm\_1, (e) Both arm reach their target location avoiding probable collision, (f) The arm again starts its trajectory for second target locations and both end-effectors get into a probable collision, (g) The arm\_2 replans its path through new path on graph structure to avoid any collision with dynamic obstacle, (h) Both arms reach second target locations safely and efficiently avoiding any collision with dynamic obstacles.}
    \label{fig:hardware1}
\end{figure*}

\subsection{Performance Analysis and Baseline Comparison}

We compare the performance of our proposed method against classical high-dimensional sampling based navigation method and state-of-the-art latent space based planning approach for obstacle avoidance with following two metrics:
\begin{itemize}
    \item \textit{Total Runtime, T(in secs)}: Total Runtime is calculated by addition of total clock time to compute the initial trajectory on latent network, total time to replan when initial path gets occluded with dynamic obstacles and total time to move each joint of robotic arm before reaching a target location.
    \item \textit{Success Ratio, SR (in percentage)}: A planned trajectory is regarded as success if the adaptive learning robotic arm efficiently reaches the target location while avoiding the dynamic obstacle created by the other arm. 
\end{itemize}

Table \ref{Table1} presents the evaluation results of our proposed approach against the baselines mentioned earlier. To ensure the thoroughness and robustness of our method, we designed two different complexity levels of planning scenarios for our dual-robot manipulation setup. In the simple complexity scenario, our robotic arms are required to reach a single random target location within the concise workspace. In contrast, the hard complexity scenario requires us to plan for multiple target locations while avoiding any potential collision effectively.
As shown in Table \ref{Table1}, our approach outperforms the baseline methods with better performance. By adopting a graph-based path planning on a pre-computed data-structure, our algorithm can efficiently find a new path through a new node to avoid collisions with non-stationary robotic arms. Compared to the other two methods, SERA demonstrates an efficient and adaptive path planning scheme, which enables faster replanning. Moreover, the success ratio comparison reveals that our method achieves a higher rate of successful trajectory completion for both complexity scenarios compared to the other two baseline methods.

%Table \ref{Table1} summarises the evaluation of our proposed approach with the mentioned baselines. To ensure thoroughness and robustness of our method, we have designed two different level of complex planning scenarios for our dual-robot manipulation setup. In simple complexity scenario, our robotic arms are required to reach single random target location respectively inside the concise workspace. On the contrary, in hard complexity scenario, the planning requires to meet condition of reaching multiple target locations while deftly avoiding any probable collision. From Table \ref{Table1}, it is apparent that our approach has outperformed the baseline approaches with better performance. Since, we approach a graph based path planning on a pre-computed data-structure, our algorithm can find a new path through a new node for avoiding collisions with non-stationary robotic arm. SERA showcases efficient and adaptive path planning scheme which enables faster replanning than other two methods. Besides, from success ratio comparison, we can easily check that for both scenarios our method has showed higher successful trajectory completion than other two baseline methods.  

\subsection{Dynamic Obstacle Avoidance}

Our research aims to develop a method based on 2D manifold representation that can concurrently avoid dynamic obstacles created by another robotic arm, namely arm\_1, and re-plan the trajectory to reach the goal position in the shortest possible time. Our approach takes advantage of the densely connected networks over the safe low-dimensional points, making it easy to perturb the initial route and form a new route by changing the intermediate node and edge connections. When the adaptive entity encounters an occlusion while following its initial trajectory, it receives sensory feedback of the possible collision and avoids the colliding nodes by re-planning the shortest path through the next best shortest path. This enables the adaptive learner to dynamically avoid obstacles while travelling from the initial pose to the goal pose. Fig.\ref{fig:adaptive} illustrates this concept, showing the adaptively modified initial routing at two different time-steps based on negative (collision) sensory feedback from the proximity sensors. Additionally, we employ the decoder part of the VAE to obtain joint poses for these selected manifold points.

% The main goal of our research work is to devise a method established on 2-D manifold representation for concurrently avoiding dynamic obstacles created by another robotic arm $arm\_1$ and re-planning its trajectory to reach goal position within shortest possible time. Since, the networks over the safe low-dimensional points are densely connected and our algorithm makes sure there exists at least one connecting edge between any two random nodes, it is effortlessly plausible to perturb the initial route and form a new route by changing the intermediate node and edge connections. Whenever the adaptive entity encounters an occlusion while following its initial trajectory, it receives a sensory feedback of possible collision and avoids following the colliding nodes. Instead the algorithm enables the robotic arm to re-plan the shortest path through the next best shortest path. Thus, the adaptive learner dynamically avoids the obstacle in between travelling to goal pose from the initial pose. This idea has been drawn in Fig.\ref{fig:adaptive} where the initial routing has adaptively modified at two different time-steps based on negative(collision) sensory feedback from the proximity sensors. Moreover, the decoder part of the VAE is employed to get joint poses for these selected manifold points. 

In Fig. \ref{fig:sim}, we present snapshots of optimized and adaptive trajectory planning in a robot-robot interaction scenario, where both arms have to maneuver in a confined workspace. In Fig. \ref{fig:sim}(b), when both robotic manipulators are in close proximity and heading towards a possible collision, proximity sensors embedded in the robot body are triggered, and the controller receives feedback to stop the current trajectory and modify the initial shortest path. The negative feedback is displayed through a red colormap in Fig. \ref{fig:sim}(b). In the next snapshot (c), we observe that arm\_2 concurrently changes its initial path to avoid a collision. We also show a red shaded image of how both arms would collide if no dynamic re-routing is engaged in our approach. Additionally, sensory feedback changes positively as the proximal distance between the two robotic entities increases.
Our algorithm achieves optimized trajectory planning and execution while adapting to dynamic obstacles. Path adaptation happens on a single node to avoid dynamic obstacles, ensuring efficient execution. In Fig. \ref{fig:sim}(d) and (e), both robotic manipulators reach their goal positions smoothly and safely.
In Fig. \ref{fig:hardware1}, we provide a sequence of robot operation in real-world settings. We show a challenging planning scenario where both robotic manipulators have to reach multiple target locations serially. The adaptive robotic arm replans through new nodes whenever the previous planned path gets occluded with dynamic obstacles created by the other robotic arm. Target locations are precomputed in the task-space of the robotic arm, and respective joint-configuration and nearest vertice representation are queried from the graph data-structure. The robot adaptively follows the path to reach the target location without getting into any collision with the other robotic arm.

\section{Conclusion}

Our research aims to develop a robotic policy that can adaptively respond to dynamically generated obstacles or state constraints. To achieve this, we propose to synergistically integrate optimally placed proximity sensors that can preemptively sense dynamic obstacles with a reactive robotic motion planning algorithm. The algorithm is based on topological manifold learning, variational autoencoder, and incremental graph traversal, which enables efficient and effective obstacle avoidance.

We believe that our proposed algorithm can have widespread applications in various empirical settings, particularly in scenarios where a global scene camera is absent, and algorithm reuse is crucial. By continuously constructing robotic policy, our algorithm can adapt to changes in the environment and achieve robust performance in dynamic situations.

% Our research aims at 
% developmentally or continuously constructing robotic policy against 
% dynamically generated obstacles or state constraints.
% Our key idea is to synergistically integrate optimally placed proximity sensors 
% that preemptively sense dynamic obstacles
% with a reactive robotic motion planning algorithm based on topological manifold learning, 
% variational autoencoder, and incremental graph traversal.
% %
% We believe that our proposed algorithm can be widely useful in many empirical settings, 
% especially when global scene camera is absent and algorithm reuse is paramount.
% 

\bibliographystyle{./bibliography/IEEEtran}
\bibliography{bib}

\end{document}